\documentclass{article}
\usepackage{ijcai24}

\usepackage{times}
\usepackage{soul}
\usepackage{url}
\usepackage[hidelinks]{hyperref}
\usepackage[utf8]{inputenc}
\usepackage[small]{caption}
\usepackage{subcaption} % 引入subcaption宏包用于创建子图

\usepackage{graphicx}
\usepackage{amsmath}
\usepackage{amsthm}
\usepackage{booktabs}
\usepackage{algorithm}
\usepackage{algorithmic}
\usepackage[switch]{lineno}
\usepackage{hyperref}
\usepackage{multirow}
\usepackage{array}
\usepackage{color}
\title{Distance-aware Attention Reshaping: Enhance Generalization of Neural Solver for Large-scale Vehicle Routing Problems}
\author{
Yang Wang$^1$
\and
Ya-Hui Jia$^1$\footnote{Corresponding Author}\and
Wei-Neng Chen$^1$\And    
Yi Mei$^2$
\affiliations
$^1$South China University of Technology, China\\
$^2$Victoria University of Wellington, New Zealand		
\emails
ftwangyang@mail.scut.edu.cn,
\{jiayahui, cschenwn\}scut.edu.cn,
yi.mei@ecs.vuw.ac.nz
}
\begin{document}

\maketitle

\begin{abstract}
   Neural solvers based on attention mechanism have demonstrated remarkable effectiveness in solving vehicle routing problems. However, in the generalization process from small scale to large scale, we find a phenomenon of the dispersion of attention scores in existing neural solvers, which leads to poor performance. To address this issue, this paper proposes a distance-aware attention reshaping method, assisting neural solvers in solving large-scale vehicle routing problems. Specifically, without the need for additional training, we utilize the Euclidean distance information between current nodes to adjust attention scores. This enables a neural solver trained on small-scale instances to make rational choices when solving a large-scale problem. Experimental results show that the proposed method significantly outperforms existing state-of-the-art neural solvers on the large-scale CVRPLib dataset. 
   %It is noteworthy that the existing NS have not demonstrated superior performance compared to the most basic traditional algorithms, particularly when dealing with large-scale VRPs. This may suggest that these NSs have inherent limitations in their design, or are not sufficiently effective in addressing broader and more complex issues.
\end{abstract}

\section{Introduction}

Vehicle Routing Problems (VRPs) are combinatorial optimization problems aimed at assigning a fleet of vehicles to a set of customers under various constraints and planning the optimal route for each vehicle to minimize the total cost~\cite{konstantakopoulos2020vehicle}. VRP is highly applicable in fields like logistics distribution, public transportation, and garbage collection~\cite{lyu2023prediction}. It is also a non-deterministic polynomial hard (NP-hard) problem, where the difficulty and time complexity of solving it increase dramatically with the scale of the problem~\cite{mor2022vehicle}. Consequently, traditional exact and heuristic algorithms often struggle to meet the demands for efficiency and quality in modern transportation and logistics~\cite{accorsi2021fast}.

\begin{figure}[t]
    \centering
    \begin{subfigure}[b]{0.65\linewidth} % Use 49% of the line width to ensure that there is some space between the two images
        \centering
        \includegraphics[width=\linewidth]{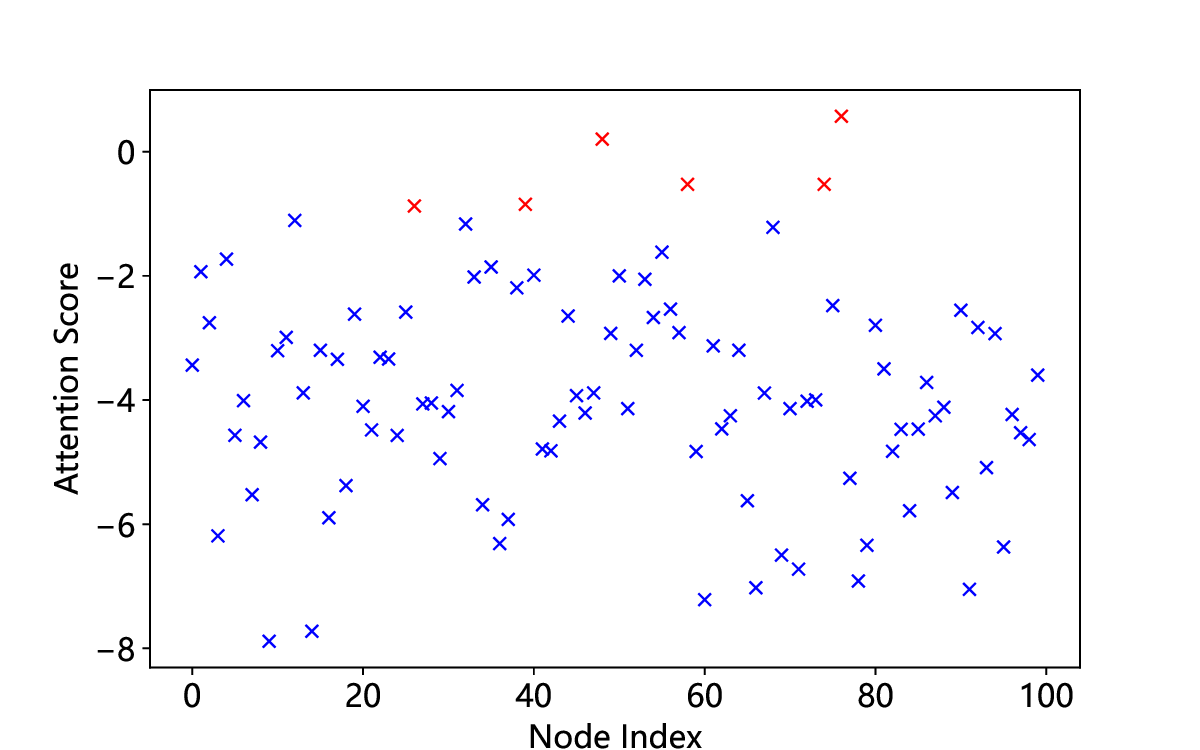}
        \caption{Attention score of 100 nodes}
        \label{fig:100}
    \end{subfigure}
    %\hfill % This command adds white space between two subgraphs
    \begin{subfigure}[b]{0.65\linewidth} % Also use 49% of the line width
        \centering
        \includegraphics[width=\linewidth]{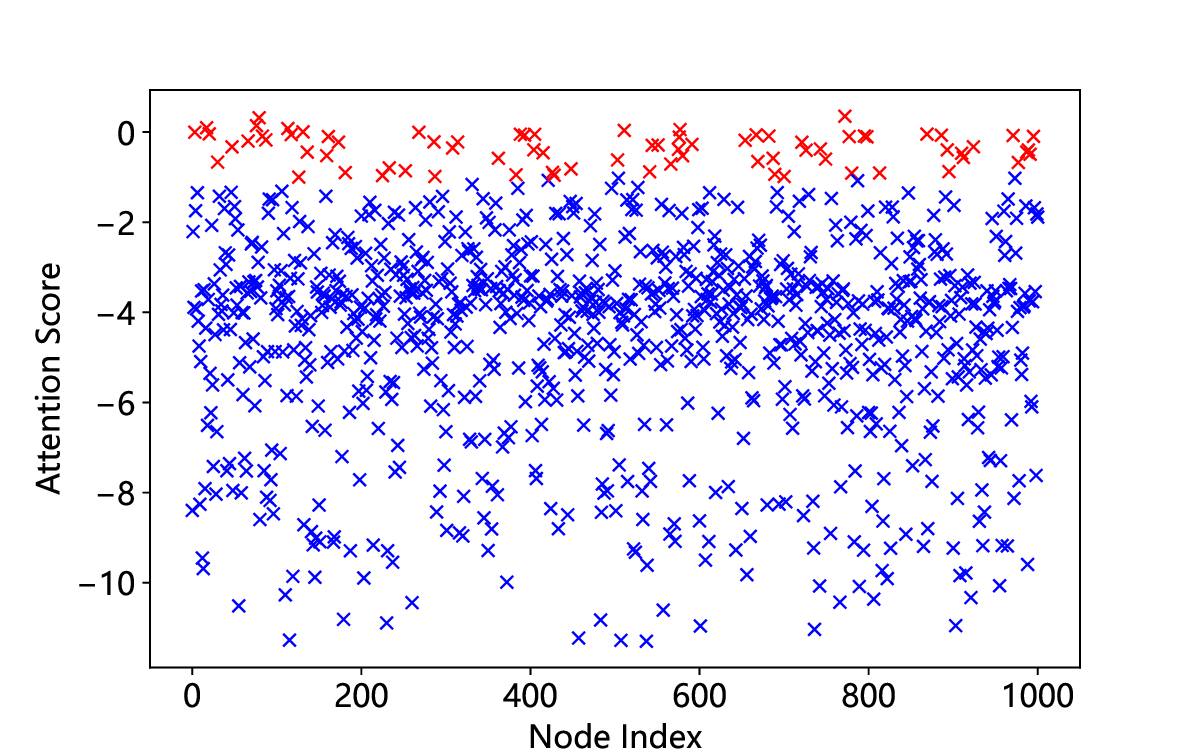}
        \caption{Attention score of 1000 nodes}
        \label{fig:1000}
    \end{subfigure}
    \caption{The distribution of attention scores when an NS trained on instances with 100 nodes tries to solve VRPs with 100 and 1000 nodes. Red points represents scores larger than -1.}
    \label{fig:1}
\end{figure}

In recent years, Neural Solver (NS), also known as Neural Combinatorial Optimization (NCO) methods, have emerged as a novel solution method and demonstrated significant performance in combinatorial optimization problems~\cite{barrett2020exploratory}. This approach leverages the powerful feature extraction capabilities of Neural Networks (NN) and the online decision-making ability of Reinforcement Learning (RL)~\cite{vinyals2015pointer,bello2016neural}. Through an end-to-end learning method, it acquires optimization strategies directly from the raw problem inputs, generating high-quality solutions. NS has shown great flexibility in adapting to different problem scenarios and data distributions, without relying on manually designed heuristic rules and optimization algorithms~\cite{hottung2021efficient,choo2022simulation}.

VRP is essentially a sequence decision problem, similar to text generation problems. Given the remarkable performance of Transformer~\cite{vaswani2017attention}~in such problems, many scholars~\cite{kool2018attention,chen2019learning}~have also applied it to solve VRPs and achieved promising results. In the solution generation process, an NS based on Transformer utilizes a self-attention mechanism to compute attention scores, which represent the importance of the currently visited customer to other unvisited customers~\cite{ma2021learning,wu2021learning}. NS selects the next customer to visit based on the attention scores, forming a visitation sequence.

%Specifically, the VRP is inherently a sequence decision problem, requiring the determination of the visiting order for a series of nodes (such as customer locations), which is remarkably similar to the sequence data types processed by the Transformer architecture, like text sequences in natural language processing. Consequently, many NS methods employ the Transformer architecture to solve the VRP. In detail, NS first uses a Transformer encoder to encode the input sequence into a hidden state sequence, containing features and relational information of the customers. Subsequently, the Transformer decoder generates the next output element, namely the next customer to visit, based on the hidden state sequence and previous output sequence. In this process, NS utilizes a self-attention mechanism to compute attention scores, which are used to represent the importance of the currently visited customer for unvisited customers. Ultimately, NS selects the next customer to visit based on the magnitude of the attention scores, forming a visitation sequence.

Currently, NCO methods have achieved promising performance on small-scale VPRs with less than 100 customers, but their performance on large-scale problems is still unsatisfactory~\cite{jiang2023ensemble}. Because of the NP-hard nature of VRP and the high complexity of the exploration mechanism of RL, directly training an NS on large-scale instances is extremely expensive. Generalizing models trained on small-scale problems to large-scale ones becomes a natural idea~\cite{fu2021generalize,drakulic2023bq}. Ideally, by doing so, we can discover learn the fundamental principles and strategies from the learned models for solving  VRPs on small-scale instances and transfer this knowledge to solving larger-scale problems~\cite{bengio2021machine}. Following this path, researchers have proposed various strategies, including decomposition, transfer,  ensemble and enhancement strategies~\cite{berto2023rl4co,liu2023good,li2023distribution,gao2023towards}. The representative state-of-the-art NS, i.e. Ensemble of Local and Global policies (ELG)~\cite{gao2023towards}, enhances the generalization ability of global NS by utilizing locally transferable topological features. Although these strategies bring some generalization ability to their NSs, their effectiveness is far from expected.

%primarily involve the fusion of data-driven and heuristic information, that is, utilizing rules, constraints, and experiential heuristics related to combinatorial optimization problems to assist NS in learning and decision-making\cite{}. 

Through our research, we have identified an interesting phenomenon in existing attention-based NS during the generalization process from small scale to large scale: the dispersion of attention scores. This dispersion implies that when solving large-scale VRPs by an NS trained on small-scale instances, the significantly increased number of customer nodes leads to a corresponding increase in high-attention-score nodes. Such a phenomenon renders the model inefficient in differentiating the importance/priority of nodes, thereby limiting the generalization ability of NS. Figures \ref{fig:1}(a) and \ref{fig:1}(b) show the distribution of attention scores when an NS trained on instances with 100 nodes tries to solve VRPs with 100 and 1000 nodes, respectively. We can see that on the 1000-node VRP instance, there are many high-attention-score nodes (red points), indicating dispersed attention scores. Under the circumstances, NS struggles to accurately pinpoint the most important nodes for the current decision.

%注意机制的内在约束:注意力机制旨在突出某些特征，同时淡化其他特征。然而，当面对大量节点时，如果未对注意力机制进行定制化的改进，这些机制可能无法有效区分众多节点的优先级，导致注意力分数分散。

%The aforementioned strategies have not resolved the issue of attention score dispersion well.
Although the aforementioned strategies can marginally affect the attention scores, they have not customized improvements to the attention scoring mechanism. Thus, they fail to effectively address the issue of attention dispersion. For example, some strategies like ELG~\cite{gao2023towards}~take human expertise as an untrained model and train the model with NN collaboratively, in which the human expertise can affect the attention scores a little. However, such a mechanism may bring the risk of diminishing the value of human experience and causing overfitting, especially when the training data fails to adequately represent the complexity and diversity of real-world problems.

Different from previous methods, in this paper, we take human expertise as a trained model. Based on this thought and the analysis of the attention score dispersion phenomenon, we propose a method called Distance-aware Attention Reshaping (DAR). The core idea of DAR is to reshape the attention scores by incorporating the Euclidean distance information between nodes, without increasing the model's complexity. This strategy encourages NS to preferentially select closer nodes in large-scale decision spaces, thereby enhancing the precision and rationality of decision-making.  Specifically, DAR incorporates heuristic information as prior knowledge within the structure of the model. This integration allows the model to directly utilize this knowledge, eliminating the necessity for further training of parameters. The contributions of this paper are summarized as follows:

\begin{itemize}

\item In exploring the generalization process from small-scale problems to large-scale problems, we discovered a phenomenon that limits the current performance of NS generalization: the dispersion of attention scores. We also analyzed two possible causes of this issue.

\item We propose a simple yet effective DAR method that reshapes the existing NS attention scores to make the attention more focused, thereby distinguishing the important nodes from massive candidates when solving large-scale VRPs.

\item Our research verifies that directly integrating expert knowledge with NS is more effective than indirectly refining expert experience through NN when NS is generalized from small to large scales.

%\item Existing NS have not surpassed the most basic greedy algorithms in terms of generalization performance from small-scale to large-scale VRP. However, our method has achieved this feat, obtaining the current state-of-the-art results in learning-based methods.

\end{itemize}

\section{Background}
In this section, we first introduce some basic concepts of VRP and how to use NS to solve a VRP. Then, we introduce the three categories of strategies for enhancing the generalization abilities of NS methods.

\subsection{Vehicle Routing Problem and Neural Combinatorial Optimization}
\textbf{Vehicle Routing Problems.} VRPs involve determining the most optimal set of vehicle routes to service a group of customers, using a fleet of vehicles with limited capacity. The objective is to ensure all customers are served while the total route length is minimized. Consider~$n$~customer nodes and~$L$~vehicles with identical capacity. Each customer node~$i$~is defined by a coordinate~$({x_i},{y_i})$~and a demand~$c_i$. Each vehicle has a capacity~$Q$. The depot node is labeled as~$0$, with coordinates~$({x_0},{y_0})$. The goal of VRP is to find a set of vehicle routes~$P = \{ {P_1},{P_2},...,{P_L}\}$~such that each customer node is visited exactly once, the load of each vehicle does not exceed its capacity, and the total cost is minimized. The mathematical formulation can be expressed as follows:
\begin{equation}
\mathop {\min }\limits_P \mathop \sum \limits_{j = 1}^L \mathop \sum \limits_{i \in {P_j}} {d_{i,i + 1}},
\end{equation}
where, $d_{i,i+1}$~is the distance between node~$i$~and node~$i$+$1$, which can be calculated using Euclidean distance, i.e. :
\begin{equation}
{d_{i,i + 1}} = \sqrt {{{({x_i} - {x_{i + 1}})}^2} + {{({y_i} - {y_{i + 1}})}^2}}.
\end{equation}

\noindent\textbf{NCO models.}
The fundamental principle of the NCO model is to create a policy network consisting of an encoder and a decoder. The encoder extracts feature vectors from the input problem, capturing the implicit relationships within the problem. The decoder makes the next decision based on the information from the encoder. Typical models include  Pointer Network~\cite{vinyals2015pointer}, Graph Neural Network~\cite{khalil2017learning}, and Transformer~\cite{bresson2021transformer}. In this paper, we select the POMO model~\cite{kwon2020pomo}~as the base NS, which is currently one of the most frequently adopted NSs in VRP studies. Since the POMO model is implemented based on the Transformer architecture, we will briefly introduce the principles related to it.

In solving a VRP, the input sequence is typically a vector composed of features such as customer coordinates and demand quantities, which means:

\begin{equation}
{\boldsymbol{h}_i} = {\rm{Encoder}}({x_i},{y_i},{c_i}),
\end{equation}
where, $\boldsymbol{h}_i$~represents the feature vector of node~$i$, and ~$x_i$,~$y_i$,~$c_i$~are the coordinates and demand of node~$i$.

Finally, an output sequence is generated through the decoder, which typically consists of a sequence of customer numbers, representing the order in which customers are visited, which means:
\begin{equation}
{\boldsymbol{a}_t} = {\rm{Decoder}}({\boldsymbol{h}_t},{s_t},\boldsymbol{h},m),
\end{equation}
where $\boldsymbol{a}_t$~is probability distribution of selecting the other nodes.~$\boldsymbol{h}_t$~represents the hidden state of the decoder. $s_t$~is the output action from the previous time step.~$\boldsymbol{h}$~denotes the output of the encoder, encompassing the feature vectors of all nodes.~$m$~is the number of trajectories of exploring nodes. Details about the POMO model can be found in ~\cite{kwon2020pomo}.

\subsection{Related Works}
\noindent\textbf{Decomposition Strategy.} This strategy aims to break down complex, large-scale problems into smaller sub-problems, simplifying the problem space and enhancing the efficiency of solutions~\cite{li2021learning,zong2022rbg,hou2022generalize,pan2023h}. In recent research,~\cite{ye2023glop}~proposed the Global and Local Optimization Policies (GLOP) framework, which employs the concept of divide-and-conquer to effectively solve a variety of large-scale VRPs. However, decomposition strategies might overlook the relationship between sub-problems. Solutions to sub-problems may not cooperate optimally with each other, hence affecting the quality of the overall problem solution.

\noindent\textbf{Transfer Strategy.} Researchers are dedicated to exploring how to utilize knowledge learned from the NCO model on one task (such as a small-scale problem) to improve its performance on another task (such as a large-scale problem)~\cite{bi2022learning,zhang2023neural}. Specifically, the methods proposed in~\cite{jiang2023multi}~and~\cite{zhou2023towards}~demonstrated that by employing comparative learning and meta-learning frameworks, it is possible to effectively extract transferable information within VRP instances. This approach yields favorable results across VRP tasks of various sizes and distributions. However, this strategy may not always ensure a smooth transfer between different problem instances, especially in cases where there are significant differences in data distribution.

\begin{figure}[t]
    \centering
    \begin{subfigure}[b]{0.49\linewidth} % Use 49% of the line width to ensure that there is some space between the two images
        \centering
        \includegraphics[width=\linewidth]{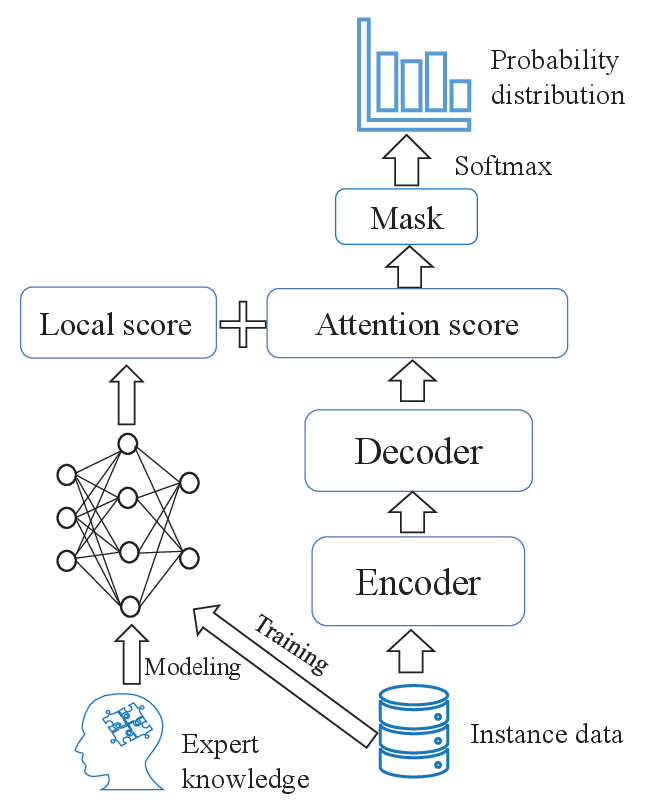}
        \caption{ELG}
        \label{ELG}
    \end{subfigure}
    \hfill % This command adds white space between two subgraphs
    \begin{subfigure}[b]{0.49\linewidth} % Also use 49% of the line width
        \centering
        \includegraphics[width=\linewidth]{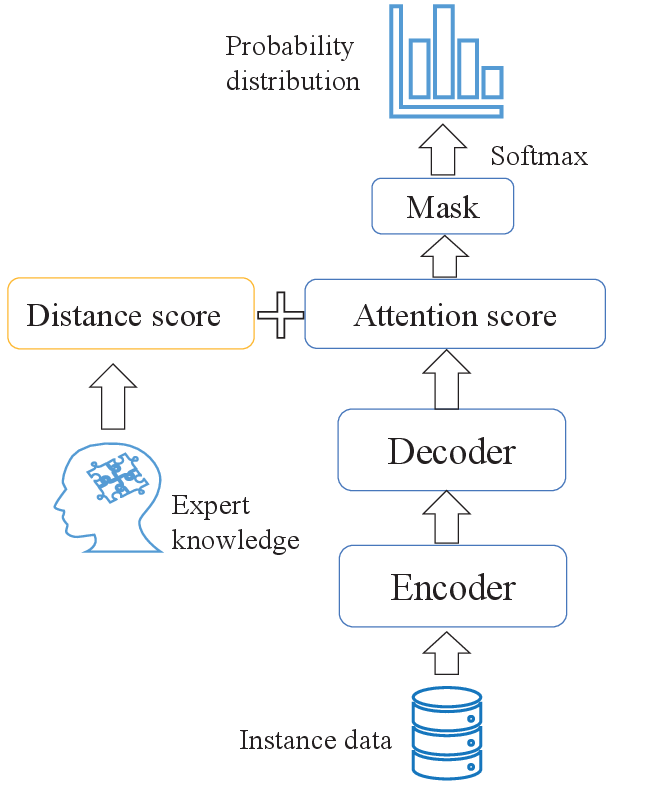}
        \caption{Our model}
        \label{DAR}
    \end{subfigure}
    \caption{The schematic diagrams of ELG and our model.}
    \label{fig:2}
\end{figure}

\noindent\textbf{Ensemble and Enhancement Strategy.} Researchers also considered enhancing the generalization performance of models for solving large-scale VRPs through combining multiple methods or enhancing existing algorithms~\cite{kim2021learning,kim2022sym,luo2023neural,jin2023pointerformer,gao2023towards}. Among these strategies, ELG, proposed in \cite{gao2023towards}, is currently the best one. To balance the exploration and exploitation, it was designed with two parts, i.e. the local topological strategies that originate from the expertise and the macro guidance of global construction strategies provided by NN, so that it can adapt to different scales of VRPs. Figure \ref{fig:2}(a) shows the schematic diagram of ELG, in which the expertise is treated as an untrained model and is trained together with NN. Although ELG has achieved very good performance, this kind of combination increases the complexity of the NS, making it more computationally expensive to train. Meanwhile, it still heavily relies on the training data, which may fail to take advantage of the expert knowledge. Thus, the generalization ability brought by expertise is limited.

%Since , these strategies often lead to increased complexity and computational demands of the model. At the same time, we speculate that an over-reliance on data-driven models might neglect or fail to fully exploit these artificial experiences, thereby limiting their ability to generalize. Figure 2 displays the fundamental distinctions between the DAR and ELG models. Without adding any extra parameters, DAR employs heuristic information as a form of prior knowledge to steer the learning and decision-making processes of NS.

\section{Proposed Method}

Different from the aforementioned methods, DAR treats expertise as a trained model. The schematic diagram of DAR is shown in Figure \ref{fig:2}(b). Before introducing DAR, the reasons behind the phenomenon of attention dispersion are analyzed in the first place. Then, the proposed method is introduced in detail.
%\subsection{Preliminary}

\subsection{Attention Dispersion Analysis}
\begin{figure}[ht!]
    \centering
    \includegraphics[width=0.5\textwidth]{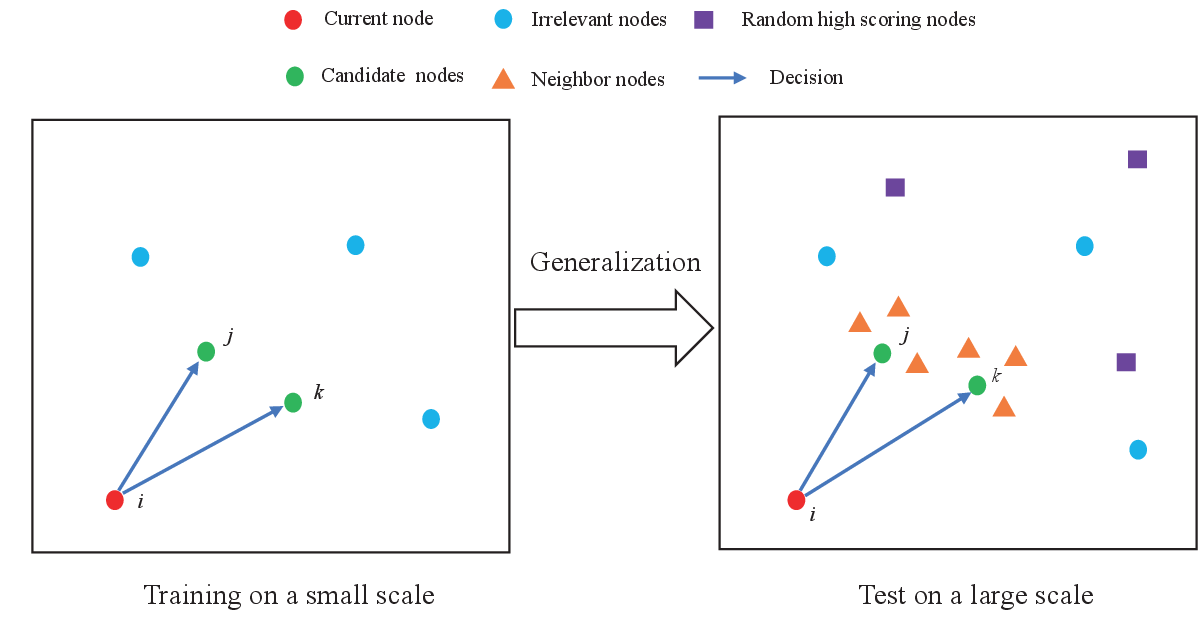} % 使用\textwidth作为图片宽度
    \caption{The emergence of attention dispersion when generalize an NS from small to large.}
    \label{fig:3}
\end{figure}

In the decoding phase, the NCO model uses an attention mechanism to assign a weight to each candidate node, indicating the probability of the node being selected. The advantage of the attention mechanism is that it can dynamically adjust the weights based on the current state and historical selections, thereby achieving adaptive decision-making. However, we have found that there is an attention dispersion phenomenon as depicted in Figure \ref{fig:1}. We think that the attention dispersion in the generalization process of NS from small to large scales comes from two perspectives: the number of neighboring nodes and the randomness of attention scores, which is illustrated in Figure \ref{fig:3}.

\textbf{(1) Number of neighboring nodes.} Assume that in small-scale instances, the current node $i$ (red point) tends to choose nodes $j$ and $k$ (green points) in the next step based on attention scores, implying that these two nodes have higher attention scores than other irrelevant nodes (blue points). When this strategy is generalized to large-scale instances, around the original candidate nodes $j$ and $k$, there might be some similar neighbors (orange triangles). Based on the semantic relationships learned by NS in small-scale instances, these neighbor nodes might also have high attention scores.

\textbf{(2) Randomness of attention scores.}  During the training process, the model can learn the semantic relationships between nodes in small-scale instances. In the process of generalizing from small to large, there will be nodes whose position or capacity information has not been captured by the trained NS, as depicted by the purple squares in Figure \ref{fig:3}. The attention scores assigned to these nodes by the model are unknown. With a certain probability, these unrelated nodes might also be given high attention scores.

%From these two aspects, we believe that in the generalization process from small to large scales, NS is likely to generate more evenly distributed attention scores, leading to attention dispersion.
%Since the weights of the attention mechanism are calculated based on the features of the input graph and the hidden state of the decoder, it may not fully utilize some heuristic information related to the problem itself, such as the distance between nodes. Therefore, the idea of utilizing this knowledge to assist NS in making wise decisions naturally emerged.

\subsection{DAR Method}

\begin{figure}[ht!]
    \centering
    \includegraphics[width=0.5\textwidth]{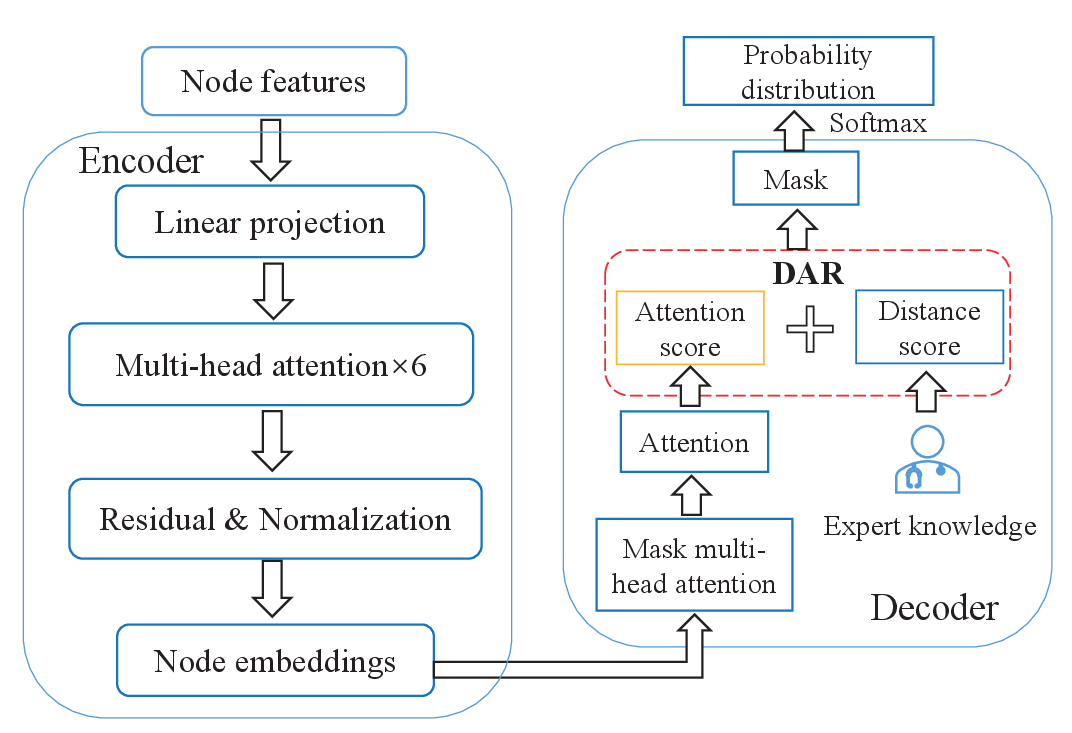} % 使用\textwidth作为图片宽度
    \caption{The pipeline of our model.}
    \label{fig:4}
\end{figure}

To address the issue of attention dispersion, we propose a DAR method that can utilize the distance information between current nodes to adjust the weights of the attention mechanism without adding extra parameters. The basic idea of our method is that for each candidate node, we calculate its distance from the currently selected node. Then, based on the magnitude of this distance, we increase or decrease its attention weight, thereby achieving a reshaping of the attention scores. The pipeline of our model is shown in Figure \ref{fig:4}.

Specifically, DAR comprises the following four steps:

\textbf{(1) Compute the attention scores for nodes.} We use the standard dot-product attention mechanism~\cite{vaswani2017attention}, where for the current state~$\boldsymbol{q}_i$ that is the query vector, we calculate its attention score as follows:
\begin{equation}
{\boldsymbol{a}_i} = \boldsymbol{q}_i^T{\boldsymbol{h}},
\end{equation}
where,~$\boldsymbol{h}$~is the feature vector output by the encoder.

\textbf{(2) Calculate the distance between the current node and other nodes.} We use Euclidean distance as the distance metric, meaning for the current node~$i$, we calculate its Euclidean distance to other nodes as:

\begin{equation}
{\boldsymbol{d}_i} = (d_{i,0}, \dots, d_{i,n}).
\end{equation}

\textbf{(3) Calculate the distance score for nodes.} Without loss of generality, we assume that all nodes are located on a square within $\left[ {0,1} \right]$$\times$$\left[ {0,1} \right]$. Thus, the maximum distance does not exceed~$\sqrt 2$. We utilize logarithmic functions to process distance values, thereby enhancing the distinction of neighboring distances. Given that optimal decisions in VRP often rely on a smaller neighborhood space around the current node~\cite{zhou2023towards}, we introduce a hyperparameter $K$ to limit the number of adjacent nodes. For nodes outside this neighborhood space, we assign negative distance scores, effectively reducing the influence of distant nodes and thus optimizing the decision-making process for selecting the next node. This can be expressed as follows:

\begin{equation}
{b_{i,j}} = \{ \begin{array}{*{20}{c}}
{- \log ({d_{i,j}}),}&{{if\rm{~}}j{\rm{~}} is{\rm{~}}the{\rm{~}}topK{\rm{~}}closest{\rm{~}}to{\rm{~}}{i}}\\
{- {d_{i,j}}}&{{otherwise\rm{}}}.
\end{array}
\end{equation}

\textbf{(4) Compute the reshaped attention scores for current nodes.} We add the attention scores and distance scores together to obtain the reshaped attention scores, which can be expressed as:

\begin{equation}
{ \boldsymbol{\tilde a}_i} = \boldsymbol{a}_i + \boldsymbol{b}_i.
\end{equation}

Then, we normalize these reshaped attention scores to obtain the final attention weights:

\begin{equation}
{\alpha _{i,j}} = \left\{ {\begin{array}{*{20}{c}}
{C \cdot \tanh ({\tilde a_{i,j}}),i \notin {\pi _{1:t - 1}}}\\
{ - \infty , otherwise}
\end{array}} \right.,
\end{equation}
where, ${\pi _{1:t - 1}}$~represents the visited nodes, and to effectively scale the results,~$C$~is set to 50~\cite{zhou2023towards}.

Finally, we use these reshaped attention weights to generate the probability distribution for the next node, thereby completing a decision-making process. According to the probability chain rule, the probability distribution of instance X can be calculated as follows:
\begin{equation}
{P_\theta }({\pi}|{X}) = softmax({\boldsymbol{\alpha} _i}) = \prod\limits_{t = 1}^n {{P_\theta }({\pi _{t} = i}|{\pi _{1:t - 1}},X)}.
\end{equation}

In summary, there are two reasons why the DAR method can alleviate attention dispersion. 

\textbf{ (1) Reduction in the number of neighboring nodes.} The DAR method amplifies the distance scores of the nearest neighbors to the current node using a logarithmic function, which guides the NS to prioritize nodes that are closer, even when the attention scores are similar. 

\textbf{(2) Identification of random high-score nodes.} DAR assists NS in excluding distant and irrelevant high-score nodes by penalizing faraway nodes.

\subsection{Reinforcement Learning Training}
To ensure fairness in the experimental results, we also adopts the classic RL algorithm REINFORCE~\cite{williams1992simple} to train our model. Specifically, rewards are given based on the quality of the solution (the negative of the total distance), and the parameters of the network are updated according to the policy gradient, which can be expressed as follows:
\begin{equation}
R(l) =  - \mathop \sum \limits_{i \in P} {c_{i,i + 1}},
\end{equation}

\begin{equation}
L(\theta ) \approx  - \frac{1}{M}\mathop \sum \limits_{m = 1}^M (R({l_m}) - \bar R)\mathop \sum \limits_{t = 1}^{n + 1} \log {p_{a_t^m,t}},
\end{equation}
where, $\theta$~represents the parameters of the network,~${p_\theta}$~is the probability distribution output by the network,~$M$~is the number of paths,~$l_m$~is the~$m$-$th$~path,~$a_t^m$~is the~$t$-$th$~action on the~$m$-$th$~path,~$n$+$1$~is the total number of actions, corresponding to visiting each of the~$n$~customer nodes and returning to the depot, and~$\bar R$~is the average of the rewards for the~$M$~paths.

\section{Experiments}
%The main objective of this section is to address the following research questions. \textbf{RQ1:} Can DAR enhance the generalization capability of NS on large-scale VRPs? \textbf{RQ2:} How is the generalization performance of the existing NS? \textbf{RQ3:} How does the DAR method alter the distribution of attention scores? \textbf{RQ4:} How does the range of the hyperparameter~$k$~affect the generalizability of the DAR method?
\subsection{Datasets}
In the training phase, this study replicates the approach of~\cite{kool2018attention}~by randomly generating node coordinates and demands for training instances. In the testing phase, we use the well-known CVRPLib\footnote{http://vrp.atd-lab.inf.puc-rio.br/index.php/en/}~dataset for case comparisons~\cite{uchoa2017new,arnold2019efficiently}. Compared to synthetic data, the CVRPLib dataset offers a range of well-documented, real-world instances, providing a fairer test for the generalization performance of our method.
\subsection{Training Setting}

In our experiments, we adopted a varying-scale training method\cite{gao2023towards}. Initially, following the training approach of previous NCO models, we trained the strategy on smaller instances ($N$ = 100). To prevent overfitting, we employed an early stopping mechanism at 200,000 steps. Subsequently, based on the small-scale training, additional training was conducted on varying-scale instances ($N$ randomly selected from $U$(100, 500)), totaling 25,000 steps. For the new parameters of DAR, we set the number of neighbor nodes ($K$) to 100. The batch size~$bs$~is adjusted based on~$N$~to prevent memory overload. It is calculated as~$bs = 120 \times {\left( {\frac{{100}}{N}} \right)^{1.6}}$. All experiments were conducted on a single NVIDIA GeForce RTX 3090 GPU equipped with 24GB of memory, using the PyTorch 1.13.0 framework built on the PyCharm platform. Other hyperparameters following the default settings of POMO~\cite{kwon2020pomo}.

\subsection{Inference Setting}
To ensure the fairness and effectiveness of the experimental results, all NCO methods only adopt a greedy strategy during the inference process and do not incorporate any additional search algorithms.
%For POMO, ELG, and GLOP, we run their source code on our test set using the default settings.
\subsection{Baselines}
We compare our method with (1) classic solvers: greedy, LKH3~\cite{helsgaun2017extension}, HGS~\cite{vidal2022hybrid}, the results of which are derived from~\cite{vidal2022hybrid}; (2) state-of-the-art NSs: baselines AM~\cite{kool2018attention}~and~POMO~\cite{kwon2020pomo}, transfer strategy Omni-VRP~\cite{gao2023towards}, decomposition strategies TAM~\cite{hou2022generalize}~and~GLOP~\cite{ye2023glop}, ensemble strategy ELG~\cite{zhou2023towards}. For greedy, POMO, and ELG, since the source codes are provided, we adopted the same training method as in this paper and ran their source code directly on the test set using the default settings. The experimental results for the remaining methods are all derived from the original paper. If the results on a data set was not provided in their original papers, they are left blank.
%\footnote{https://github.com/wouterkool/attention-learn-to-route}
%\footnote{https://github.com/yd-kwon/POMO}\footnote{https://github.com/CIAM-Group/NCO\_code}
%\footnote{https://github.com/RoyalSkye/Omni-VRP}\footnote{https://github.com/gaocrr/ELG}
%\footnote{https://github.com/henry-yeh/GLOP} LEHD~\cite{luo2023neural}, 
\subsection{Generalization Performance on CVRPLib}%(RQ1 \& RQ2)

In this section, we initially selected 100 instances from the CVRPLib Set-X for testing, with each instance having a size between 100 and 1,000. Then, we evaluated the generalization performance of NS on 35 expanded Set-XML instances from~\cite{gao2023towards}, with instance sizes $N$ ranging from 100 to 5,000.  Further, we extended our analysis to include 6 large-scale CVRPLib Set-XXL instances:  $Antwarp1$, $Antwarp2$, $Leuven1$, $Leuven2$, $Ghent1$ and $Ghent2$. These instances significantly vary in scale, with the number of nodes ($N$) ranging from 3,000 to 11,000, and feature complex node distributions, offering a comprehensive view of the models' performance across different scales and complexities. 

Due to differences in strategies for enhancing generalization performance,  training methods of NS and the sizes of problem instances it solves also differ. Table \ref{tab:1} shows the gap comparison between the DAR method and the state-of-the-art transfer strategy and ensemble strategy with the Best Known Solutions (BKS) under the same training environment. Table \ref{tab:2} presents the gap comparison between the DAR method and state-of-the-art decomposition strategies with the BKS, across 32 Set-X instances and 6 Set-XXL instances.

\noindent\textbf{Overall performance. } The experimental results show that on 100 Set-X, 35 Set-XML, and 6 Set-XXL instances, DAR consistently achieved the lowest gap, at 5.27\%, 5.20\%, and 10.82\%, respectively. This demonstrates the effectiveness of the DAR method in enhancing the generalization performance of NS from small scale to large scale. Traditional solvers struggle with large-scale instances due to prolonged computation times, especially with over 10,000 nodes. Therefore, no comparison was made with traditional algorithms such as LKH3 and HGS on instances ranging from 3,000 to 11,000 in scale. To our knowledge, our method currently represents state-of-the-art performance in solving large-scale VRPs.

\noindent\textbf{Heuristics VS DAR. } Compared to the basic greedy algorithm, on the Set-X instances, DAR reduced the gap from 21.03\% to 5.66\%, on the Set-XML instances from 19.64\% to 5.20\%, and on the Set-XXL instances from 13.08\% to 10.82\%. This demonstrates that besides the distance score based on our human expertise, the knowledge learned through NS learning still plays a key role in the generalization process, and these two parts cooperate well.
Although the DAR method did not achieve the performance of LKH3 and HGS  on Set-X and Set-XML instances, when facing new instances, the DAR method can effectively utilize GPU for parallel processing, which is particularly efficient for solving large-scale VRPs. In contrast, on Set-X instances, LKH-3 and HGS need to run for at least 16 minutes on an Intel Gold 6148 Skylake 2.4 GHz processor to achieve the aforementioned performance, while DAR only requires 0.39 seconds. 

\noindent\textbf{Baselines VS DAR. } Compared to the baselines, DAR significantly outperforms AM and POMO. On Set-XXL instances, both AM and POMO exhibit noticeable performance degradation, with gaps exceeding 30\%, while DAR remains stable at around 10\%. This suggests that distance scores can effectively guide NS in large-scale generalization.

\noindent\textbf{Transfer strategy VS DAR.} Compared to the state-of-the-art transfer strategy Omni-VRP, DAR shows improvements, indicating that the rational application of expert knowledge has more potential than knowledge transfer. 

\noindent\textbf{Ensemble strategy VS DAR.} Compared to the state-of-the-art integration strategy ELG, DAR outperforms ELG on Set-X instances (5.66\% VS 5.92\%), on Set-XML instances (5.20\% VS 5.95\%), and on Set-XXL instances (10.82\% VS 16.08\%). This indicates that directly using expert knowledge is more effective than learning expert knowledge in the process of generalizing from small to large scale.

\noindent\textbf{Decomposition strategy VS DAR.} On Set-XXL instances, compared to the latest decomposition strategies like TAM, TAM-AM, GLOP, and GLOP-LKH3, DAR's generalization performance does not show a significant decline and is advantageous among various strategies, demonstrating the robustness of the DAR method.

\begin{table}[htbp]\scriptsize
\centering
\setlength{\tabcolsep}{2pt} % Adjust the space between columns
\begin{tabular}{ccccc}
\toprule
& \multirow{2}{*}{Method} & \multicolumn{1}{c}{Set-X(100 instances)} & \multicolumn{1}{c}{Set-XML(35 instances)} & \multicolumn{1}{c}{Set-XXL(6 instances)} \\
& &0.1$k<N<$1$k$ & 0.1$k<N<$5$k$ &3$k<N<$11$k$ \\
\midrule

& BKS & 0.00\% &0.00\% & 0.00\%  \\
\midrule
\multirow{3}{*}{\rotatebox{90}{Heu}}
& Greedy & 21.03\% &19.64\% & 13.08\%  \\
& LKH3 & 1.00\% &- & -  \\
& HGS & 0.11\% &0.00\% & -  \\
\midrule
\multirow{4}{*}{\rotatebox{90}{NS}}
%& AM & - &- & 48.15\%  \\
& POMO & 8.31\% &10.51\% & 30.85\%  \\
%& LEHD & 12.52\%  &- & - \\
& Omni-VRP & 6.47\% &10.60\% & -  \\
%& TAM-AM & - & -& 30.07\%  \\
%& TAM-LKH3 & -  &-& 22.50\%  \\
%& GLOP & - &- & 19.75\%  \\
%& GLOP-LKH3 & - &- & 18.80\%  \\
& ELG & 5.92\% &5.95\% & 16.08\%  \\
& DAR & \textbf{5.66\%} & \textbf{5.20\%}& \textbf{10.82\%}  \\
\bottomrule
\end{tabular}
\caption{Comparisons between DAR and greedy, LKH3, HGS, POMO, Omni-VRP, ELG on Set-X, Set-XML, and Set-XXL.}
\label{tab:1}
\end{table}

\begin{table}[htbp]\footnotesize
\centering
\setlength{\tabcolsep}{2pt} % Adjust the space between columns
\begin{tabular}{cccc}
\toprule
& \multirow{2}{*}{Method} & \multicolumn{1}{c}{Set-X(32 instances)}  & \multicolumn{1}{c}{Set-XXL(6 instances)} \\
& &0.5$k<N<$1$k$  &3$k<N<$11$k$ \\
\midrule

& BKS & 0.00\%  & 0.00\%  \\
\midrule
\multirow{6}{*}{\rotatebox{90}{NS}}
& AM &24.65\% & 48.15\%  \\
& TAM-AM &10.97\% & 30.07\%  \\
& TAM-LKH3   &9.87\%& 22.50\%  \\
& GLOP  &- & 19.75\%  \\
& GLOP-LKH3&- & 18.80\%  \\
& DAR & \textbf{5.27\%} & \textbf{10.82\%}  \\
\bottomrule
\end{tabular}
\caption{Comparisons between DAR and AM, TAM-AM, TAM-LKH3, GLOP, GLOP-LKH3 on Set-X and Set-XXL.}
\label{tab:2}
\end{table}

\subsection{Generalized performance details of CVRPLib Set-XXL}

\begin{table*}[htbp]\scriptsize
\centering
\setlength{\tabcolsep}{3pt} % Adjust the space between columns
%\begin{tabular}{lcc|cc|cc|cc}
\begin{tabular}{cccccccccccccccccc}
\toprule
\multirow{2}{*}{Instance} & \multirow{2}{*}{Scale} & \multirow{2}{*}{BKS} & \multicolumn{3}{c}{Greedy} & \multicolumn{3}{c}{POMO}  & \multicolumn{3}{c}{ELG} & \multicolumn{3}{c}{GLOP} & \multicolumn{3}{c}{DAR}\\
\cmidrule(lr){4-6} \cmidrule(lr){7-9} \cmidrule(lr){10-12} \cmidrule(lr){13-15} \cmidrule(lr){16-18}
 &  &  & Cost & Gap & Time & Cost & Gap & Time & Cost & Gap & Time& Cost & Gap & Time& Cost & Gap & Time\\
\midrule
$Leuven1$ & 3$k$ & 192848 &210855 &9.34\% &2s &227782  &18.11\% &7s  & 210672 & 9.24\% & 8s & 225439 & 16.9\% & 2s &206866& 7.27\% & 7s \\
$Leuven2$ & 4$k$ & 111395  &133237 &19.61\% &3s &153489  &37.79\% &11s  & 130863 & 17.48\% & 13s& 135679 & 21.8\%  & 3s&129282 & 16.06\% & 12s \\
$Antwerp1$ & 6$k$ & 477277 &522605 &9.49\% &7s &537165 &12.55\% &26s  & 513806 & 7.65\% & 31s & 573924 & 20.3\% & 3s &509886 & 6.83\% & 28s \\
$Antwerp2$ & 7$k$ & 291350  &336561 &15.52\% &9s &378942  &30.06\% &34s  & 333543 & 14.48\% & 40s & 347871 & 19.4\% & 4s &327887 & 12.54\% & 36s \\
$Ghent1$ & 10$k$ & 469531 &508384 &8.27\% &18s &720055 &53.36\% &79s  & 553702 & 18.69\% & 84s & 564846 & 20.3\% & 5s&504399  & 7.43\% & 81s \\
$Ghent2$ & 11$k$ & 257748 &299687 &16.27\% &22s &343374  &33.22\% &86s  & 333517 & 29.40\% & 99s & 308782 & 19.8\% & 6s&295858  & 14.79\% & 90s \\

\bottomrule
\multicolumn{1}{c}{Average} & \multicolumn{1}{c}{} & \multicolumn{1}{c}{299858}& \multicolumn{1}{c}{335221} & \multicolumn{1}{c}{13.08\%} & \multicolumn{1}{c}{10s} & \multicolumn{1}{c}{393468} & \multicolumn{1}{c}{30.85\%} & \multicolumn{1}{c}{41s}   & \multicolumn{1}{c}{346184} & \multicolumn{1}{c}{16.08\%} & \multicolumn{1}{c}{46s} & \multicolumn{1}{c}{359080} & \multicolumn{1}{c}{19.75\%} & \multicolumn{1}{c}{\textbf{4s}} & \multicolumn{1}{c}{\textbf{329029}} & \multicolumn{1}{c}{\textbf{10.82\%}} &
\multicolumn{1}{c}{42s}\
 \\
\bottomrule
\end{tabular}
\caption{Detailed comparisons between DAR and greedy, POMO, ELG, GLOP on Set-XXL.}
\label{tab:3}
\end{table*}

To better understand the advantages of DAR and the limitations of existing NS, comparative experiments were set up against the greedy algorithm, baseline POMO, ensemble strategy ELG, and decomposition strategy GLOP. Table \ref{tab:3} displays the experimental details for different cases, including cost, gap, and time. The results show that DAR can effectively generalize to 6 different large-scale instances.

\noindent\textbf{Greedy VS DAR. } The DAR method outperforms the greedy algorithm (10.82\% VS 13.08\%), indicating that the knowledge learned by NS is effective, and the decision process is collaboratively accomplished by distance scores and attention scores. Apart from the DAR method, the existing NS does not surpass the traditional greedy algorithm in terms of performance on 6 large-scale Set-XXL instances, highlighting the significant limitations of the current NS. Notably, the ELG method, despite utilizing the concept of greediness, performs worse than the greedy algorithm (16.08\% VS 13.08\%), suggesting that the knowledge refined by NN might reduce the performance of the original expert knowledge.

\noindent\textbf{POMO VS DAR. } The POMO model exhibited a noticeable decline in generalization performance on instances like $Leuven2$, $Antwerp2$, and $Ghent1$, with gaps exceeding 30\%. In contrast, our method showed stable performance across  6 large-scale Set-XXL instances. This suggests that the ability of NS to learn rules from small-scale instances and automatically generalize them to large-scale instances is limited, making the direct application of expert experience and prior knowledge crucial.

\noindent\textbf{ELG VS DAR. } In the cases of $Ghent1$ and $Ghent2$, the gap for ELG was found to be 18.69\% and 29.40\%, respectively, compared to DAR's 7.43\% and 14.79\%, indicating a significant performance degradation. This suggests that direct application of expert knowledge may more easily yield generalization from small to large scale compared to learning methods. Furthermore, as DAR does not add any additional network parameters, its average inference time on 6 large-scale Set-XXL instances is also faster than that of ELG.

\noindent\textbf{GLOP VS DAR. } GLOP, through its decomposition strategy, divides large-scale problems into smaller ones, each solved using NS. Although this reduces the solving time, it is prone to getting trapped in local optima. This results in an average gap of 19.75\% on 6 large-scale Set-XXL instances, compared to only 10.82\% for DAR.

\subsection{DAR Visualization} 

\begin{figure}[t]
    \centering
    \begin{subfigure}{0.49\linewidth}
        \centering
        \includegraphics[width=\linewidth]{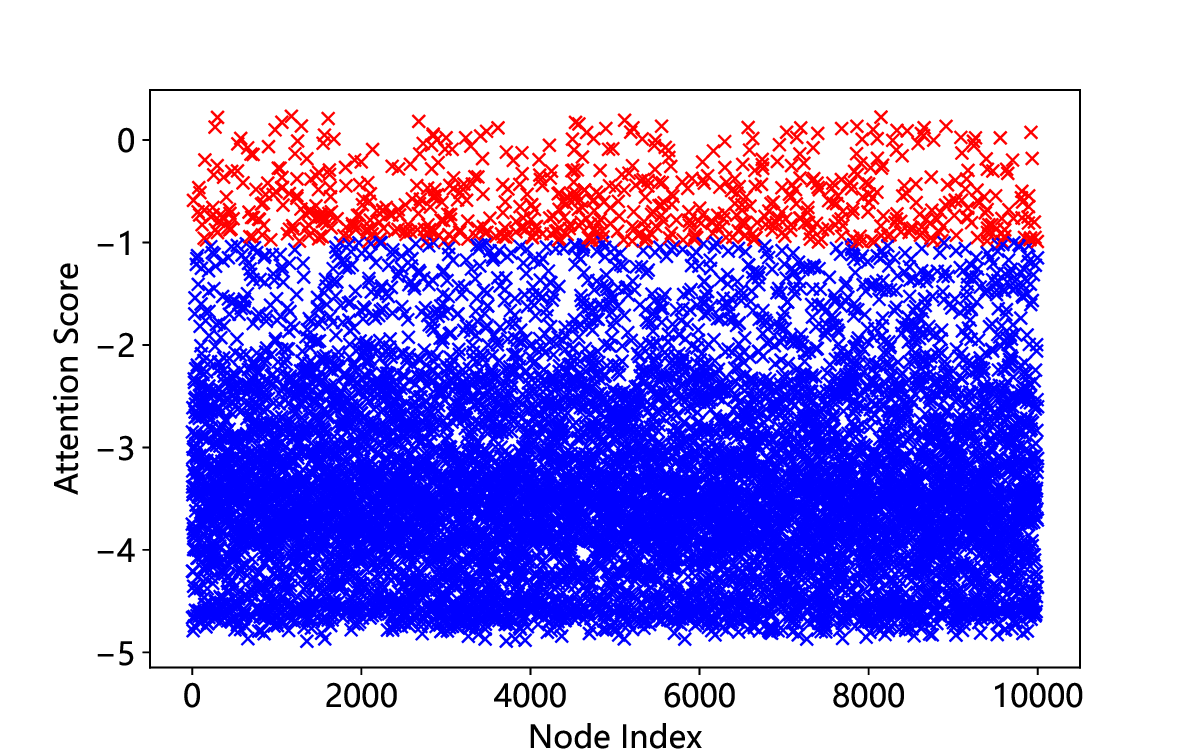}
        \caption{POMO attention score}
        \label{fig:51}
    \end{subfigure}
    \hfill % 添加这一行来分隔两个并排的子图
    \begin{subfigure}{0.49\linewidth}
        \centering
        \includegraphics[width=\linewidth]{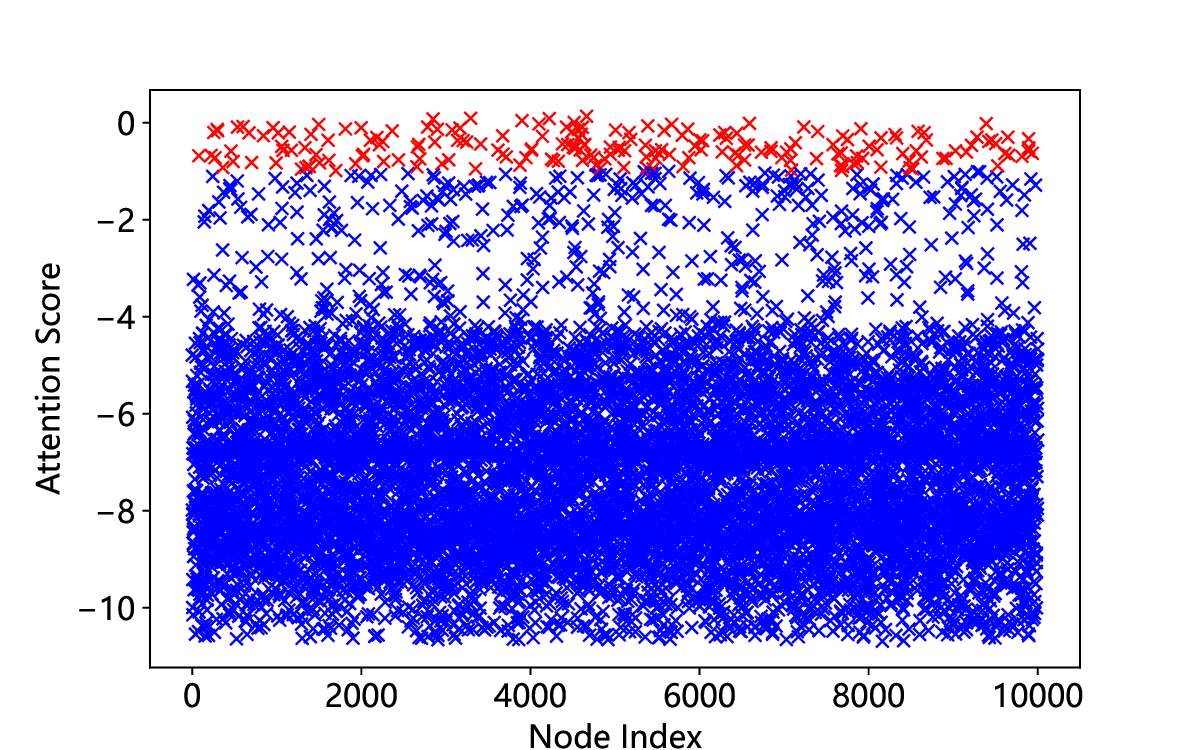}
        \caption{ELG attention score}
        \label{fig:52}
    \end{subfigure}\\ % 第一行结束
    \begin{subfigure}{0.5\linewidth}
        \centering
        \includegraphics[width=\linewidth]{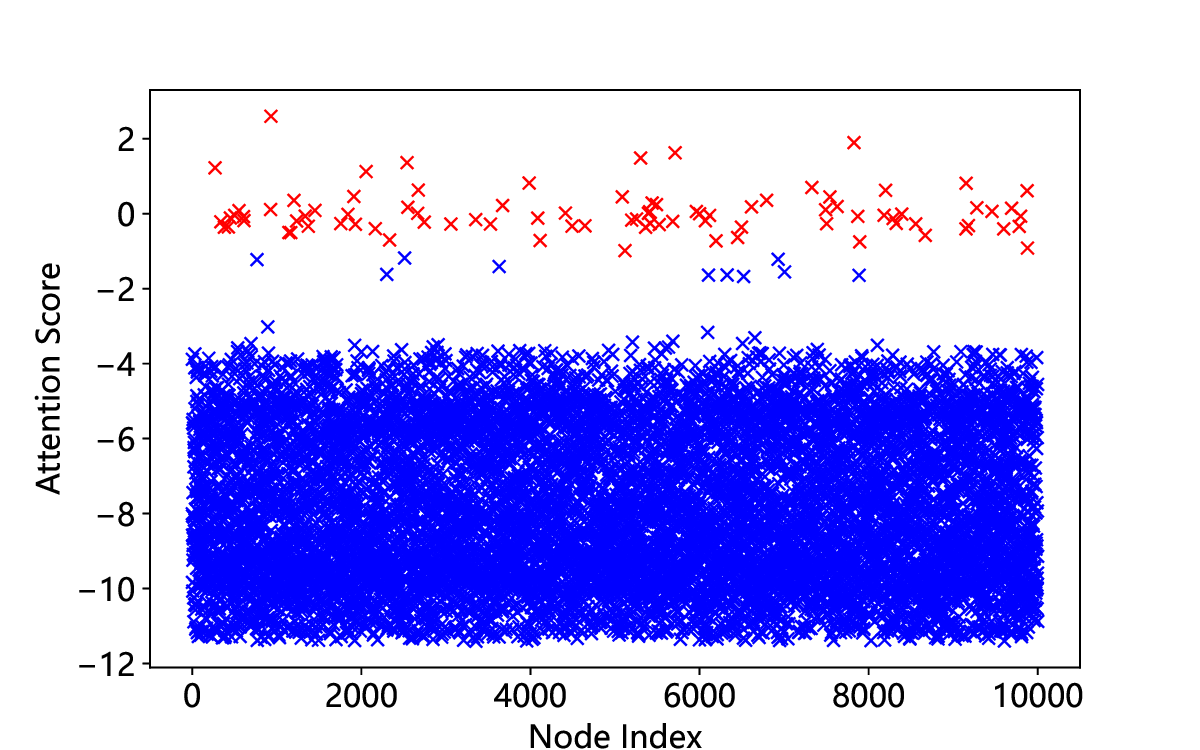}
        \caption{DAR attention score}
        \label{fig:53}
    \end{subfigure}
    \caption{Case Ghent1 attention score.} 
    \label{fig:5}
\end{figure}

To more intuitively demonstrate the impact of the DAR method on the attention scores of NS, we conducted a visual analysis on a specific VRP instance. Specifically, we chose the $Ghent1$ instance from CVRPLib, which consists of 10,000 customer nodes and one depot node. 

Figure \ref{fig:5} displays the distribution of attention scores during the first step decision-making of POMO, ELG, and DAR on the given instance. From Figure \ref{fig:5}(a), it can be seen that the POMO model assigns a higher number of high scores to all candidate nodes at current time step, as indicated by the red points. This suggests that the POMO model struggles to distinguish the importance of different nodes, leading to the selection of incorrect or inefficient nodes. Figure \ref{fig:5}(b) shows the ELG attention scores, which, although reduced in some high-scoring nodes compared to POMO, have a limited effect, and the phenomenon of dispersed attention still exists.  In contrast, Figure \ref{fig:5}(c) demonstrates that DAR allocates attention scores with clear differences to candidate nodes at each time step, alleviating the issue of dispersed attention, as indicated by the reduced number of red high-score nodes. This implies that DAR can utilize distance information to adjust attention scores, enabling NS to make more rational choices in large-scale decision spaces.

 %\noindent\textbf{Discussion.} Overall, by observing the attention score distribution of distance scores and DAR scores, it can be indirectly demonstrated that the DAR strategy enhances the performance of NS. Specifically, DAR improves the attention scores of each node by leveraging distance information, leading NS to preferentially select nodes closer to the current node. This approach effectively reduces the likelihood of choosing incorrect or inefficient nodes, making the selection process more precise and efficient.

\subsection{Ablation Study}%(RQ4)

Table \ref{tab:4} illustrates the impact of the hyperparameter $K$ at various values (20, 50, 100) on the generalization of large-scale problems. The results indicate that as the value of $K$ increases, there is an improvement in generalization performance. By setting the number of neighbor nodes to 100, DAR can achieve an effective balance between local and global information during the generalization process from small scale to large scale in NS.

\begin{table}[h!]\footnotesize
\centering
%\begin{tabular}{@{}lccc@{}}
\begin{tabular}{cccc}
\toprule
Instance & $K$ = 20 & $K$ = 50 & $K$ = 100 \\
\midrule

$Leuven1$    & 7.99\% & 6.84\% & 7.27\% \\
$Leuven2$    & 21.05\% & 16.17\% & 16.06\% \\
$Antwerp1$   & 8.09\% & 6.92\% & 6.83\% \\
$Antwerp2$   & 26.89\% & 13.41\% & 12.54\% \\
$Ghent1$     & 13.76\% & 7.73\% & 7.43\% \\
$Ghent2$     & 41.04\% & 17.63\% & 14.79\% \\
Set-X total avg.& 6.60\% & 6.01\% & 5.66\% \\
Set-XML avg. & 7.85\% & 7.18\% & 5.20\% \\
Set-XXL avg. & 19.79\% & 11.62\% & 10.82\% \\

\bottomrule
\end{tabular}
\caption{Empirical results of different neighbor node sizes.}
\label{tab:4}
\end{table}

\section{Conclusion}
In this paper, we investigate the generalization ability of NS in solving VRPs from small to large scales and identifies a key issue: the dispersion of attention scores. In light of this, we analyze two reasons leading to this issue and propose a DAR method to mitigate it. Experiments on the CVRPLib dataset show that the DAR method significantly outperforms existing NSs. Additionally, we find that despite the complex and advanced structures of NSs, their performance in large-scale practical applications is not always superior to the basic greedy algorithm. This suggests that directly utilizing expert knowledge and prior information is a simple yet effective way to enhance the generalization performance of NSs, especially in facing large-scale and complex VRPs.

In future research, we aim to further enhance the generalization ability of NSs from two aspects. First, advanced training methods or encoding scheme should be studied to solve the dispersion of attention score in essence. Moreover, we will consider other effective ways to incorporate more expert knowledge or heuristics.

%\section*{Acknowledgements}

%Acknowledgements are optional.

\bibliographystyle{named}
\bibliography{ijcai24}

\end{document}